\documentclass[11pt]{article}
\usepackage[final]{acl}
\usepackage{times}
\usepackage{latexsym}
\usepackage[T1]{fontenc}
\usepackage[utf8]{inputenc}
\usepackage{microtype}
\usepackage{inconsolata}
\usepackage{graphicx}
\usepackage{enumitem}
\usepackage{booktabs}
\usepackage{lineno}
\usepackage{caption}
\usepackage{subcaption}
\usepackage{adjustbox}
\usepackage{dirtytalk}
\usepackage{url}
\usepackage{pgfplots}
\pgfplotsset{compat=1.18}
\usepackage{pgf-pie}
\usepackage{amssymb}
\usepackage{multirow}
\usepackage{soul}
\usepackage{changepage}
\usepackage[most]{tcolorbox} 
\usepackage{tabularx} 

\usepackage{array}
\newcommand{\pheadL}{\toprule
  & \multicolumn{2}{c}{Full} & \multicolumn{2}{c}{LoRA} \\
  \cmidrule(lr){2-3}\cmidrule(lr){4-5}
  & 135M & 360M & 135M & 360M \\\midrule}
\newcommand{\phead}{\toprule
  \multicolumn{2}{c}{Full} & \multicolumn{2}{c}{LoRA} \\
  \cmidrule(lr){1-2}\cmidrule(lr){3-4}
  135M & 360M & 135M & 360M \\\midrule}
\newcommand{\ptabL}[1]{{\footnotesize\setlength{\tabcolsep}{3.5pt}%
  \begin{tabular}{w{l}{5.6em}*{4}{w{c}{2.6em}}}\pheadL #1\bottomrule\end{tabular}}}
\newcommand{\ptab}[1]{{\footnotesize\setlength{\tabcolsep}{3.5pt}%
  \begin{tabular}{*{4}{w{c}{2.6em}}}\phead #1\bottomrule\end{tabular}}}

\title{Family Matters: Language Transfer and Merging\\ for Adapting Small LLMs to Faroese} 

\author{Jenny Kunz \\
  Linköping University \\
  \texttt{jenny.kunz@liu.se} \\\And
  Iben Nyholm Debess \\
  University of the  Faroe Islands  \\
  \texttt{ibennd@setur.fo}  \\\And
  Annika Simonsen \\
  University of Iceland \\
  \texttt{ans72@hi.is} \\}

\begin{document}
\maketitle
\begin{abstract}
In this paper, we investigate strategies for adapting small language models to Faroese, a low-resource North Germanic language. Starting from English-pretrained models, we first continue pre-training on other Scandinavian languages, either individually or combined via model merging or multilingual pre-training, before fine-tuning on Faroese. 
To address the lack of existing Faroese evaluation resources, we construct two new minimal-pair probing benchmarks, one for linguistic acceptability and one for text comprehension, and complement them with human evaluations conducted by native Faroese linguists. 
Our results show that transfer from related languages improves language modeling and linguistic acceptability, while gains on comprehension probes are inconsistent across model sizes and tuning methods.
Icelandic is the strongest single source language for linguistic acceptability, but no single language is reliably best for comprehension. 
Under full fine-tuning, merging improves on the best single source language in perplexity, outperforms joint multilingual pre-training, and gives the best comprehension scores. With LoRA however, joint multilingual pre-training is the stronger initialization. 
\end{abstract}

\section{Introduction}
\label{sec:introduction}

Using LLMs in low-resource languages is still challenging: model quality is tightly linked to data availability \cite{robinson-etal-2023-chatgpt, li2024languagerankermetricquantifying}, and even basic comprehension may fail \cite{court-elsner-2024-shortcomings}. Adequate coverage of these languages typically requires the largest available multilingual models, if it works at all. Therefore, in this paper, we focus on specializing small, efficient models in individual languages. 

We focus on Faroese, a North Germanic language spoken by approximately 70,000 people, primarily in the Faroe Islands. Training data is scarce, with only 95 million words available in the deduplicated FineWeb-2 dataset \cite{penedo2024FineWeb-2}, which makes it worthwhile to transfer linguistic knowledge from related languages. Transfer for Faroese could draw from multiple sources: descended from Old Norse and shaped by a distinctive historical and sociopolitical trajectory, it has been described as \textit{the central Nordic language} \cite{torp_nordiske_1998}, sharing structural features with all other Scandinavian languages to varying degrees. 

We investigate transfer strategies for adapting small generative models to Faroese through continued pre-training (CPT). Prior work has shown that syntactic similarity is a strong predictor of transfer success \cite{chang-etal-2024-multilinguality}, and that Faroese encoder models benefit from Icelandic and other Scandinavian languages \cite{snaebjarnarson-etal-2023-transfer}. Beyond single-language transfer, we explore parameter merging \citep{pmlr-v162-wortsman22a, ilharco2023editing, yadav2023tiesmerging} as a means of combining models trained on different languages. This allows fine-grained control over the relative influence of each source language---a flexibility that would be costly to achieve through multilingual CPT---and enables us to balance the closer morphological and lexical proximity of Icelandic against the syntactic similarity and larger data resources of the mainland Scandinavian languages. 

A persistent challenge for low-resource languages is the scarcity of evaluation data. We address this in two ways. First, we introduce two minimal-pair evaluation suites: FoBLiMP, targeting linguistic acceptability, and FoBCoMP, targeting text comprehension, both combining existing, newly collected, and adapted datasets. 
Second, we conduct expert human evaluations by native Faroese linguists on two generation tasks, one with and one without downstream fine-tuning. 

Our results confirm that transfer from related languages improves language modeling and linguistic acceptability, whereas gains on comprehension probes are inconsistent. Icelandic transfer yields the strongest performance on linguistic probes but no source language is consistently best for comprehension. Merging shows promise under full fine-tuning, where merged models (slightly) improve the best single-language models in perplexity and outperform joint multilingual CPT, and are the strongest models on comprehension for both sizes under full fine-tuning. For parameter-efficient CPT with LoRA \citep{hu2022lora} however, multilingual CPT outperforms merging. 

\section{Background and Related Work}
\label{sec:background}

\paragraph{Faroese Typology} 
Faroese belongs to the Insular Scandinavian branch together with Icelandic, while Norwegian, Swedish, and Danish form Mainland Scandinavian. This classification reflects differences in lexicon, morphology, and syntax, although Faroese also shares substantial traits with the Mainland languages \cite{thrainsson_faroese_2012}. 
Lexically, Faroese shares cognates with both Icelandic and Danish, though no quantitative study exists \cite{jacobsen_foroysk_2021,jacobsen2022faroese}. Morphologically, it is closest to Icelandic, with some overlap with Norwegian \cite{torp_nordiske_1998}. Syntactically, however, Faroese aligns more closely with Mainland Scandinavian languages \cite{ussery2023ditransitives,debess_en_2017,petersen_dynamics,petersenheycock2017,sandoy2005icelandicfaroese}. Faroese therefore occupies an intermediate position: morphologically and lexically closest to Icelandic, but syntactically closer to the Mainland languages.

To quantify Faroese’s position, we compute lexical and syntactic similarity measures. Faroese is not represented in major typological databases and lacks representations in multilingual toolkits such as lang2vec \citep{littell-etal-2017-uriel, wals}, therefore we rely on alternative resources. 
\begin{table}[t]
\centering
\begin{tabular}{lcc}
\toprule
\textbf{Language} & \textbf{Lexical} & \textbf{Syntactic} \\ \midrule
Danish             & 0.167 & \textbf{0.814} \\
Icelandic          & \textbf{0.321} & 0.803 \\
Norwegian, Bokmål  & 0.180 & 0.812 \\
Norwegian, Nynorsk & 0.205 & 0.790 \\
Swedish            & 0.181 & 0.794 \\ \bottomrule
\end{tabular}
\caption{Lexical and Syntactic Similarity of Faroese to other Scandinavian languages. \textit{Lex. Sim.:} arithmetic mean of N-gram overlap and normalized Levenshtein similarity. \textit{Synt. Sim.:} arithmetic mean of POS-tag distribution overlap, POS-tag N-gram overlap, and POS-sequence normalized Levenshtein similarity.
}
\label{tab:faroese-similarity}
\end{table}
For \textit{lexical} similarity, we use the pan-Nordic dictionary Islex \cite{ulfarsdottir-2014-islex}. We compute normalized Levenshtein distance and character N-gram overlap between Faroese lemmas and their equivalents in other Scandinavian languages; the overall score is the unweighted mean of these measures; results are given in Table~\ref{tab:faroese-similarity}.
While this metric does not explicitly model cognate relations or orthographic differences, it reproduces the hierarchy suggested in the linguistic literature and confirms Icelandic as lexically closest.
For \textit{syntactic} similarity, we use the parallel splits of Flores-200 \cite{costa-jussa_no_2022}, POS-tagged using Stanza \cite{qi-etal-2020-stanza}. We compute three tag-based metrics: global POS distribution overlap, POS N-gram overlap, and normalized Levenshtein similarity on POS sequences; the final score is their unweighted mean. All languages show high syntactic similarity to Faroese, with Danish being the closest. 
The results support existing linguistic descriptions: Icelandic is morphologically and lexically closest to Faroese, while Mainland Scandinavian languages show stronger syntactic similarity. This typological profile motivates our experimental design, where we consider all Scandinavian languages, with particular emphasis on Icelandic. 

\paragraph{Transfer Languages}
Selecting effective transfer languages is generally crucial for language adaptation. \citet{chronopoulou-etal-2023-language} show that sharing adapter parameters across related languages improves translation for low-resource languages, and \citet{faisal-anastasopoulos-2022-phylogeny} find that linguistically informed adapter designs benefit unseen languages. In translation, syntactic similarity is predictive of transfer performance \citep{lee-etal-2022-pre}, although transfer success may depend more on source corpus size and subword overlap than on broader linguistic similarity \citep{lin-etal-2019-choosing}. 

\paragraph{Merging} 

Model merging provides an alternative to multi-task training by combining several fine-tuned models into a single checkpoint, often improving generalization.
In simple linear merging \citep{pmlr-v162-wortsman22a, choshen2022fusingfinetunedmodelsbetter}, the parameters of the models are averaged using scalar weights $\alpha_i$ that determine how much each model influences the final result. 
Task Arithmetic \citep{ilharco2023editing} instead operates on task vectors, defined as the difference between a fine-tuned model and a shared base model. These vectors represent the updates introduced by each task. The merged model is formed by adding a weighted combination of these task vectors back to the base model, again with weight coefficients $\alpha_i$.
TIES \citep{yadav2023tiesmerging} further reduces interference between tasks. First, it introduces a density hyperparameter $p \in (0,1]$ that determines how many parameters from each task vector are retained. 
Second, TIES resolves sign conflicts across models. For each parameter, it determines the dominant update direction across tasks and enforces this consensus direction. 
The final merged model is obtained by adding the weighted, sparsified, and sign-aligned task vectors back to the base model.

For task fine-tuning, merged models often preserve base model performance and have been shown to generalize better to unseen tasks; merging models fine-tuned on multiple tasks separately matches or exceeds joint multi-task training \citep{yadav2024mattersmodelmergingscale}. 
The effect of merging \textit{languages} rather than \textit{tasks} is however less explored. \citet{aakanksha2024mix} show that models safety-aligned on one language at a time and then merged outperform data mixing. For continued pre-training however, \citet{glocker2025growmergescalingstrategies} find that merged models' performance is lower than the performance of the monolingual base models. 
A key challenge, especially in merging LoRA models, is however \textit{parameter interference}, where updates from different models conflict in the shared parameter space \citep{stoica2024modelmergingsvdtie, tang2024parameterefficient, zhao2025merging, zhang-zhou-2025-unraveling}.

\section{Experimental Setup}
\label{sec:experimental_setup}

\subsection{Training}
\label{secsec:training}

We use the two smaller SmolLM2 \citep{allal2025smollm2smolgoesbig} models (135M and 360M parameters) as they are fully open, including their training data, and well-trained for their size on an English corpus. Although their size limits performance on knowledge-intensive tasks, it allows us to continually pre-train and compare them in different setups across substantial corpora. 
We experiment with two adaptation setups: full-parameter fine-tuning and parameter-efficient LoRA fine-tuning \citep{hu2022lora}.
We train for 5 epochs on the Faroese corpus (following \citet{muennighoff2023scaling}'s scaling law for data-constraint training); we do not repeat data for other languages. 
Training details can be found in Table~\ref{tab:hyperparams} in Appendix~\ref{app:a}.
We train on the deduplicated FineWeb-2 \citep{penedo2024FineWeb-2} portions for the Scandinavian languages, containing 27B tokens for Danish, 25B for Swedish, 30B for Norwegian-Bokmål, 1.6B for Icelandic, 495M for Norwegian-Nynorsk, and 95M for Faroese. Due to resource constraints, we limit the corpora for Swedish, Danish and Norwegian (Bokmål) to 4B tokens. 
We perform sequential CPT, first on an individual transfer language, then on Faroese, the latter with 3 random seeds. We do not merge the data of the source languages with the Faroese data because this may introduce language confusion and language mixing in model outputs \citep{li2025rethinkingmultilingualcontinualpretraining}. In addition to the single-language models, we train a multilingual CPT baseline on the combined Swedish, Danish, Bokmål, Nynorsk and Icelandic data, also followed by CPT on Faroese\footnote{All models and benchmarks are available in \href{https://huggingface.co/collections/jekunz/adaptation-of-smollm-to-faroese}{this HuggingFace collection}. The code is available on \href{https://github.com/jekunz/family-matters}{GitHub}.}.

\subsection{Merging} 
\label{secsec:expsetup_merging}

We apply TIES merging to the CPT models in different languages, using Mergekit \citep{goddard-etal-2024-arcees}: starting from SmolLM, we fine-tune on each language to obtain \textit{language vectors}, merge them, and train the resulting model on Faroese. 
We select three promising merges that cover different language mixes and weightings:
\textbf{Merge$^{eq}$} where we merge all five models equally (with pre-normalization weight 1 and density 0.5)\footnote{This results in a bias towards the mainland Scandinavian languages as four of them are included in the merge.}; \textbf{Merge$^{is+}$} where we merge with bias towards Icelandic (weight 1 for Icelandic, 0.5 for all others; density 0.5), and \textbf{Merge$^{da+is}$}, where we merge only two models: Icelandic and Danish (both with weight 1; density 0.5), as Danish is one of the mainland Scandinavian languages with the lowest perplexity after Faroese CPT (see Table~\ref{tab:post_adaptation_perplexities_main}). 

\subsection{Automatic Evaluation}
\label{subsec:autoeval}

We evaluate the perplexity on the validation set of the Faroese portion of FineWeb-2. 
In addition, we introduce two benchmarks: FoBLiMP for linguistic acceptability probes, and FoBCoMP for text comprehension probes. We report results on the original SmolLM models as well as for the much larger multilingual Apertus-8B model \citep{hernandez-cano-etal-2026-apertus} as baselines. 

\paragraph{FoBLiMP}
To probe zero-shot linguistic skills, we use minimal pairs with one correct and one corrupted sentence, measuring the percentage of times the model assigns higher probability to the correct sentence. This collection is called FoBLiMP (Faroese Benchmark of Linguistic Minimal Pairs). Table~\ref{tab:foblimp_overview} in Appendix~\ref{app:dataset_statistics} provides an overview of sources, subsets and modifications, and statistics.
To evaluate \textbf{subject-verb agreement}, we use the Faroese portion of MultiBLiMP \citep{jumelet2025multiblimp10massivelymultilingual}, containing 232 sentences. 
ScaLA \citep{nielsen-2023-scandeval} contains sentences corrupted by \textbf{swapping or deleting words}. Originally a binary classification task, we convert it to minimal pairs by realigning correct and incorrect sentences using Levenshtein distance ($\geq0.85$), with unmatched samples added manually. Concatenating all subsets gives 552 pairs for \textit{flip\_neighbors} and 601 pairs for \textit{delete}.
GermDetect \citep{michael-horbach-2025-germdetect} provide automatically corrupted sentences with \textbf{verb placement errors}. After removing pairs with no corruption, we obtain 2,026 pairs. As Faroese allows flexible word order, some corruptions are grammatical, but we conclude from an inspection that the original sentence is mostly more common.
We also construct minimal pairs from a human evaluation in \citet{scalvini-etal-2025-rethinking}, where two raters annotated \textbf{errors in English-to-Faroese translations} from four models. We pair translations with an error difference of at least 2, keeping those with no more than four errors in the better translation and excluding translations containing foreign scripts. This yields 680 pairs.

\paragraph{FoBCoMP}
Evaluating small LLMs in text comprehension is particularly challenging because evaluations often mix formal competence (e.g., grammar) with functional competence (e.g., following prompts) \citep{kydlicek2024finetasksmultilingualtasks}. Limited fine-tuning data further complicates comparisons. To address this, we also use text comprehension probes in a minimal-pair format. We introduce a set of five probes, called FoBCoMP (Faroese Benchmark of Text Comprehension Minimal Pairs). Table~\ref{tab:fobcomp_overview} in Appendix~\ref{app:dataset_statistics} provides an overview with additional statistics. 
We adapt the Faroese \textbf{news sentiment} dataset \citep{debess-etal-2024-good} (original labels: positive, negative, neutral) into minimal pairs by adding a sentiment-bearing sentence (\say{Hetta er gott/ringt}). Neutral labels are excluded as initial experiments showed that words such as \say{neutral} are never the most probable choice. We evaluate sentence- and article-level samples, keeping only items annotators of the original dataset agreed on, resulting in 91 sentence-level (55 positive, 36 negative) and 84 article-level (51 positive, 33 negative) pairs.
Using the same dataset, we filter GPT-4-assigned \textbf{topic labels} confirmed by a human. Minimal pairs consist of one correct topic and one incorrect topic (not assigned to the article), with related-topic pairs curated to make the task realistic (e.g., \textit{Local News} vs. \textit{International News}). This yields 234 topic classification pairs.
We also adapt the \textbf{extractive QA} dataset FoQA \citep{simonsen-etal-2025-foqa} (2,000 Faroese question–context–answer triplets) into minimal pairs via two methods: \textbf{(1) Dataset Shuffling:} Replacing the correct answer passage with an incorrect but plausible passage from another sample within the context, matching token length. This creates 21,867 pairs. \textbf{(2) GPT-4 Adversarial Answers:} Generating one alternative incorrect answer per sample that is also a span in the dataset, matching token length when possible. Exact length matches occurred for 611 answers; deviations were an average of 1.69 tokens longer than the correct answers. This yields 2,000 pairs.

\subsection{Human Evaluation}
\label{subsec:humeval}

We conduct a human evaluation to assess outputs in their linguistic quality, naturalness, and contextual appropriateness.
Given the resource-intensive nature of expert evaluation, we focus on four 360M models representing the best-performing configurations, varying along two dimensions: full fine-tuning versus LoRA, and transfer from Icelandic versus Merge$^{is+}$.
Two native Faroese speakers serve as evaluators, both trained linguists with extensive NLP evaluation experience.
We first followed a structured calibration protocol: evaluators jointly examined sample outputs, discussed edge cases, and iteratively refined the scoring guidelines until reaching consensus. Following this calibration phase, evaluators independently annotated a small set of approximately 10 samples, then met to resolve disagreements and clarify ambiguities before proceeding independently. Periodic check-ins were held throughout to maintain consistency without sharing individual scores. 

\paragraph{Sentence Continuation}
Given the limited capacity of small base models, we adopt sentence continuation as a straightforward generation task. Models are prompted with sentences drawn from a small manually compiled corpus of academic papers and local news articles (not included in FineWeb-2), with the final words removed and a trailing space added. All models produce running text, enabling direct comparison of output quality. A token cut-off of 100 was applied.
Outputs are evaluated along four subdimensions of linguistic quality, each scored 0--5: \textbf{lexical correctness} (valid Faroese vocabulary, absence of hallucinated forms), \textbf{grammatical accuracy} (morphological and syntactic well-formedness, including spelling and typography), \textbf{semantic coherence} (meaningful, internally consistent content), and \textbf{fluency/naturalness} (native-like expression). Each annotator evaluated 400 continuations (100 prompts $\times$ 4 models). 
During guideline development, annotators agreed to rate the full generated output rather than the immediate completion alone, as the truncated sentence was typically resolved within a few words. 
Inter-annotator agreement, measured by Pearson's $r$, was 0.546 overall (see Table~\ref{tab:iaa} in Appendix~\ref{app:iaa}). 

\paragraph{Summarization} 
Initial zero-shot experiments (i.e., prompting models directly to summarize a given text) failed to produce recognizable summaries, necessitating fine-tuning. As there was no Faroese summarization dataset, we constructed a synthetic one: 151 authentic texts spanning the same domains as the evaluation set (academic, news, blog), each paired with a summary generated by Claude Sonnet 4.\footnote{Fine-tuning details are provided in Table~\ref{tab:hyperparams} in Appendix~\ref{app:a}.}
For evaluation, we selected 50 source texts not in FineWeb-2 and generated summaries from all four models (token cutoff: 400), yielding 200 summary--source pairs. Annotators rated two criteria: \textbf{task completion} (the extent to which the model successfully performed the summarization task) and \textbf{linguistic quality}, each rated on a 0--5 scale. Guidelines were developed through the same calibration process as for sentence continuation. 
Inter-annotator agreement was substantially higher than for sentence continuation (Pearson's r=0.879 overall; Table~\ref{tab:iaa} in Appendix~\ref{app:iaa}), likely because of the coarser-grained criteria. 

\section{Results and Discussion}
\label{sec:results_discussion}

\begin{table*}[ht]
\centering
\begin{subtable}[t]{0.38\textwidth}
\centering
\ptabL{
En & 95.11 & 96.26 & 98.99 & 97.99 \\\midrule
+Da & 95.26 & 95.26 & 98.56 & 97.99 \\
+Is & 96.84 & \textbf{97.70} & 98.99 & 97.99 \\
+No$^B$ & 95.98 & 96.70 & 97.27 & \textbf{98.28} \\
+No$^N$ & 94.68 & 96.12 & 97.84 & 97.41 \\
+Sv & 96.70 & 95.98 & 97.70 & 98.13 \\\midrule
Merge$^{eq}$ & \textbf{98.13} & 96.55 & 97.41 & 97.41 \\
Merge$^{is+}$ & 96.55 & 96.41 & \textbf{99.14} & 98.13 \\
Merge$^{da+is}$ & 97.13 & 96.41 & 97.99 & 98.13 \\\midrule
Multilingual & 94.97 & 93.97 & 97.84 & 96.41 \\
}
\caption{Subject-verb agreement (MultiBLiMP).\\Baseline: 66.81 (135M), 70.69 (360M). \\ Apertus-8B: 91.81. Median SD: 0.43}
\label{tab:multiblimp}
\end{subtable}\hfill
\begin{subtable}[t]{0.29\textwidth}
\centering
\ptab{
91.79 & 93.96 & 94.99 & 96.20 \\\midrule
94.57 & 94.44 & \textbf{96.32} & 96.20 \\
93.06 & 94.87 & 95.47 & 96.86 \\
93.24 & 94.26 & 95.05 & 96.38 \\
93.00 & \textbf{94.99} & 94.81 & 95.23 \\
\textbf{95.17} & 94.93 & 95.89 & 95.59 \\\midrule
94.69 & 94.26 & 95.95 & 95.65 \\
94.81 & 94.50 & 95.47 & \textbf{96.92} \\
93.96 & 94.93 & 94.93 & 96.74 \\\midrule
93.12 & 93.84 & 96.07 & \textbf{96.92} \\
}
\caption{ScaLA: \emph{flip\_neighbors}.\\BL: 59.96 (135M), 62.50 (360M). \\ Apertus-8B: 90.40. Median SD: 0.34}
\label{tab:flip}
\end{subtable}\hfill
\begin{subtable}[t]{0.29\textwidth}
\centering
\ptab{
86.80 & 90.63 & 91.13 & 94.12 \\\midrule
89.63 & 91.57 & 92.24 & 94.68 \\
88.96 & \textbf{93.29} & 92.24 & 95.56 \\
89.74 & 90.90 & 92.96 & 95.56 \\
88.80 & 91.63 & 92.68 & 93.68 \\
89.02 & 92.07 & 92.85 & 95.45 \\\midrule
89.02 & 91.63 & 91.96 & 94.07 \\
\textbf{89.96} & 92.79 & 92.29 & 94.84 \\
89.41 & 93.23 & 91.79 & 94.12 \\\midrule
86.86 & 92.29 & \textbf{93.40} & \textbf{96.12} \\
}
\caption{ScaLA: \emph{delete}.\\BL: 65.39 (135M), 67.39 (360M). \\ Apertus-8B: 94.84. Median SD: 0.34}
\label{tab:delete}
\end{subtable}

\medskip

\begin{subtable}[t]{0.38\textwidth}
\centering
\ptabL{
En & 92.76 & 94.36 & 95.00 & \textbf{96.02} \\\midrule
+Da & 93.67 & 94.72 & 96.02 & 95.87 \\
+Is & 94.34 & 95.79 & 95.95 & 95.84 \\
+No$^B$ & \textbf{94.87} & 95.28 & 95.54 & 95.80 \\
+No$^N$ & 92.97 & 94.80 & 94.87 & 95.33 \\
+Sv & 93.96 & 95.26 & 95.80 & 95.92 \\\midrule
Merge$^{eq}$ & 93.95 & 95.41 & \textbf{96.07} & 95.61 \\
Merge$^{is+}$ & 94.42 & 95.79 & 95.43 & 95.79 \\
Merge$^{da+is}$ & 93.75 & \textbf{96.02} & 95.54 & 95.95 \\\midrule
Multilingual & 93.62 & 94.92 & 95.64 & 95.77 \\
}
\caption{Verb placement (GermDetect).\\Baseline: 52.22 (135M), 57.95 (360M). \\ Apertus-8B: 89.80. Median SD: 0.19}
\label{tab:germdetect}
\end{subtable}\hfill
\begin{subtable}[t]{0.29\textwidth}
\centering
\ptab{
69.95 & 70.69 & 75.59 & 75.74 \\\midrule
72.01 & 75.15 & 76.47 & 76.91 \\
71.47 & 75.78 & 77.21 & \textbf{79.46} \\
72.25 & 74.17 & 76.57 & 78.48 \\
72.35 & 73.14 & 74.61 & 74.85 \\
\textbf{73.33} & 75.00 & \textbf{77.99} & 77.84 \\\midrule
72.94 & 75.64 & 76.76 & 75.69 \\
72.50 & 74.85 & 77.25 & 75.25 \\
71.72 & \textbf{76.13} & 76.13 & 76.47 \\\midrule
69.41 & 72.60 & 77.30 & 79.17 \\
}
\caption{Translation Pairs.\\BL: 42.65 (135M), 46.32 (360M). \\ Apertus-8B: 69.20. Median SD: 0.43}
\label{tab:transl_pairs}
\end{subtable}\hfill
\begin{subtable}[t]{0.29\textwidth}
\centering
\ptab{
87.28 & 89.18 & 91.14 & 92.01 \\\midrule
89.03 & 90.23 & 91.92 & 92.33 \\
88.93 & \textbf{91.49} & 91.97 & \textbf{93.14} \\
89.21 & 90.26 & 91.48 & 92.90 \\
88.36 & 90.13 & 90.96 & 91.30 \\
89.64 & 90.65 & 92.05 & 92.59 \\\midrule
\textbf{89.74} & 90.70 & 91.63 & 91.68 \\
89.65 & 90.87 & 91.92 & 92.19 \\
89.19 & 91.34 & 91.28 & 92.28 \\\midrule
87.59 & 89.52 & \textbf{92.05} & 92.88 \\
}
\caption{Mean of the individual FoBLiMP scores in \ref{tab:multiblimp}--\ref{tab:transl_pairs}. BL: 57.41 (135M), 60.97 (360M). Apertus-8B: 87.21}
\label{tab:mean_foblimp}
\end{subtable}

\caption{Linguistic probes on datasets included in FoBLiMP: Percentage of samples where a higher probability was assigned to the original than to the corrupted sample. Results are means over 3 seeds.}
\label{tab:foblimp}
\end{table*}

\subsection{Benchmarks}

We first discuss the benchmarks results with respect to their difficulty and reliability. 
\textbf{FoBLiMP} results (Table~\ref{tab:foblimp}) show that models perform well on most probes, suggesting these tasks are relatively easy. The exception is \textit{Translation Pairs}, where scores are lower\footnote{This may be partly because translation error counts do not always reflect linguistic quality: translation errors can also reflect aspects such as incorrect content compared to the source sentence. Our annotators' judgment is, however, that most samples do contain language errors. }. 
\textbf{FoBCoMP} (Table~\ref{tab:fobcomp}) shows lower and more mixed scores than FoBLiMP, reflecting the tasks’ difficulty for small models.  
In sentiment analysis, 135M LoRA models perform poorly, with little improvement over the base model, and full fine-tuning shows similar limitations at the article level, indicating 135M models are too small for zero-shot comprehension. 
Topic classification results are challenging to interpret due to small, variable data; individual results should generally be interpreted cautiously. 
For extractive QA, results vary based on the setup. On the shuffled dataset, transfer offers no consistent gains. On the harder GPT-4–picked answers, transfer improves scores consistently for the larger 360M models. 

\textbf{Apertus-8B} has lower scores than all adapted models on the FoBLiMP mean and all FoBLiMP tasks except ScaLA \emph{delete}. On FoBCoMP, it slightly exceeds our best model thanks to strong extractive QA scores, while our 360M models are better at topic classification and sentiment analysis.

\begin{table*}[ht]
\centering
\begin{subtable}[t]{0.38\textwidth}
\centering
\ptabL{
En & 60.81 & 61.54 & 62.27 & 67.03 \\\midrule
+Da & 66.30 & 79.12 & 61.90 & 73.63 \\
+Is & 75.82 & 75.09 & 60.44 & \textbf{79.85} \\
+No$^B$ & 63.74 & 69.23 & 63.74 & 73.99 \\
+No$^N$ & \textbf{85.35} & 73.63 & \textbf{69.60} & 68.50 \\
+Sv & 60.81 & 71.43 & 64.84 & 66.67 \\\midrule
Merge$^{eq}$ & 76.56 & 83.15 & 60.44 & 63.74 \\
Merge$^{is+}$ & 63.00 & \textbf{84.25} & 60.44 & 67.77 \\
Merge$^{da+is}$ & 74.36 & 77.29 & 60.44 & 67.03 \\\midrule
Multilingual & 67.03 & 69.96 & 60.44 & 71.79 \\
}
\caption{Binary sentiment analysis (Sentences).\\Baseline: 60.44 (135M), 60.44 (360M). \\ Apertus-8B: 62.64. Median SD: 1.10}
\label{tab:sentence-pos-neg}
\end{subtable}\hfill
\begin{subtable}[t]{0.29\textwidth}
\centering
\ptab{
60.71 & 61.90 & 61.90 & 65.48 \\\midrule
61.51 & \textbf{76.19} & 61.90 & 69.05 \\
\textbf{67.46} & 73.41 & 60.71 & \textbf{71.03} \\
61.51 & 69.44 & 61.90 & 67.46 \\
\textbf{67.46} & 65.87 & \textbf{62.30} & 62.70 \\
61.90 & 64.29 & 59.92 & 61.51 \\\midrule
66.27 & 71.03 & 60.71 & 63.10 \\
59.92 & 72.22 & 60.71 & 67.46 \\
64.68 & 73.02 & 60.71 & 62.30 \\\midrule
62.70 & 73.81 & 60.71 & 64.68 \\
}
\caption{Bin.\ sentiment (Articles).\\BL: 59.52 (135M), 60.71 (360M). \\ Apertus-8B: 70.24. Median SD: 0.69}
\label{tab:articles-pos-neg}
\end{subtable}\hfill
\begin{subtable}[t]{0.29\textwidth}
\centering
\ptab{
72.22 & 77.35 & 68.66 & 76.92 \\\midrule
75.21 & 66.95 & 70.66 & 72.79 \\
73.79 & 76.07 & 73.08 & 71.37 \\
69.94 & 79.77 & \textbf{74.22} & \textbf{79.63} \\
53.56 & 66.95 & 61.25 & 75.78 \\
61.97 & 74.64 & 73.22 & 74.07 \\\midrule
66.52 & 79.20 & 61.11 & 72.22 \\
72.36 & \textbf{80.06} & 60.54 & 72.22 \\
\textbf{76.92} & 66.81 & 65.95 & 78.63 \\\midrule
63.11 & 77.07 & 63.11 & 79.34 \\
}
\caption{Topic Classification (Articles).\\BL: 54.70 (135M), 60.26 (360M). \\ Apertus-8B: 58.55. Median SD: 1.10}
\label{tab:topic_classification}
\end{subtable}

\medskip

\begin{subtable}[t]{0.38\textwidth}
\centering
\ptabL{
En & 69.67 & 78.89 & \textbf{67.00} & \textbf{84.15} \\\midrule
+Da & 66.71 & \textbf{84.20} & 65.93 & 77.00 \\
+Is & \textbf{72.53} & 78.80 & 64.00 & 81.88 \\
+No$^B$ & 68.94 & 82.38 & 63.10 & 82.48 \\
+No$^N$ & 67.09 & 82.49 & 65.01 & 83.19 \\
+Sv & 67.36 & 80.23 & 61.45 & 77.93 \\\midrule
Merge$^{eq}$ & 60.61 & 81.01 & 63.16 & 74.96 \\
Merge$^{is+}$ & 65.26 & 81.73 & 64.98 & 78.24 \\
Merge$^{da+is}$ & 69.32 & 83.76 & 64.85 & 79.35 \\\midrule
Multilingual & 71.17 & 80.31 & 64.55 & 76.91 \\
}
\caption{Extractive QA (Shuffled DS).\\Baseline: 63.06 (135M), 68.95 (360M). \\ Apertus-8B: 95.40. Median SD: 0.49}
\label{tab:extractive-qa-shuffled}
\end{subtable}\hfill
\begin{subtable}[t]{0.29\textwidth}
\centering
\ptab{
52.40 & 51.77 & 60.53 & 63.60 \\\midrule
55.77 & 65.13 & \textbf{64.85} & 70.40 \\
47.63 & 63.23 & 56.88 & 72.80 \\
56.83 & 60.18 & 57.42 & 67.13 \\
50.78 & 62.35 & 55.52 & 68.75 \\
\textbf{58.37} & 62.07 & 56.32 & 70.75 \\\midrule
55.43 & 65.27 & 57.35 & 65.57 \\
55.52 & 64.92 & 55.57 & 68.47 \\
57.18 & \textbf{67.77} & 62.77 & 71.87 \\\midrule
51.73 & 62.57 & 62.58 & \textbf{73.32} \\
}
\caption{Extractive QA (LLM gen.).\\BL: 32.90 (135M), 35.45 (360M). \\ Apertus-8B: 97.40. Median SD: 0.43}
\label{tab:extractive-qa-gpt}
\end{subtable}\hfill
\begin{subtable}[t]{0.29\textwidth}
\centering
\ptab{
63.16 & 66.29 & 64.07 & 71.44 \\\midrule
65.10 & 74.32 & \textbf{65.05} & 72.57 \\
67.45 & 73.32 & 63.02 & \textbf{75.39} \\
64.19 & 72.20 & 64.07 & 74.14 \\
64.85 & 70.26 & 62.74 & 71.78 \\
62.08 & 70.53 & 63.15 & 70.19 \\\midrule
65.08 & 75.93 & 60.56 & 67.92 \\
63.21 & \textbf{76.64} & 60.45 & 70.83 \\
\textbf{68.49} & 73.73 & 62.94 & 71.84 \\\midrule
63.15 & 72.74 & 62.28 & 73.21 \\
}
\caption{Mean of the individual FoBCoMP scores in \ref{tab:sentence-pos-neg}--\ref{tab:extractive-qa-gpt}. BL: 54.12 (135M), 57.16 (360M). Apertus-8B: 76.84}
\label{tab:focomp-mean}
\end{subtable}
    \caption{Text Comprehension Probes on datasets included in the FoBCoMP benchmark. Percentage of samples where a higher probability was assigned to the correct than to the incorrect sample. Means over 3 seeds. }
    \label{tab:fobcomp}
\end{table*}

\subsection{Transfer Languages are Important}
\label{secsec:rq1}

\begin{table}[ht]
    \centering
    \small
        \adjustbox{max width=\linewidth}{
        \begin{tabular}{lcccc}\toprule
        &\multicolumn{2}{c}{Full} & \multicolumn{2}{c}{LoRA} \\
        \cmidrule(lr){2-3}\cmidrule(lr){4-5}
        & 135M & 360M & 135M & 360M \\\midrule
En & 5.26 & 4.20 & 4.21 & 3.54 \\
\midrule
+Da & 4.51 & 3.74 & 3.74 & 3.19 \\
+Is & 4.62 & 3.69 & 3.93 & 3.21 \\
+No$^B$ & 4.51 & 3.89 & 3.76 & 3.20 \\
+No$^N$ & 4.80 & 3.91 & 4.04 & 3.45 \\
+Sv & 4.54 & 3.74 & 3.72 & 3.19 \\
\midrule
Merge$^{eq}$ & 4.50 & 3.65 & 4.02 & 3.46 \\
Merge$^{is+}$ & 4.49 & \textbf{3.63} & 3.96 & 3.37 \\
Merge$^{da+is}$ & \textbf{4.49} & 3.64 & 3.97 & 3.40 \\
\midrule
Multilingual & 5.02 & 4.01 & \textbf{3.63} & \textbf{3.11} \\\bottomrule
        \end{tabular}
        }
    \caption{Average per-token perplexity on the FineWeb-2 evaluation set after CPT on Faroese. Mean over 3 seeds.}
    \label{tab:post_adaptation_perplexities_main}
\end{table}

\begin{figure}[t]
    \centering
    \includegraphics[width=0.5\textwidth]{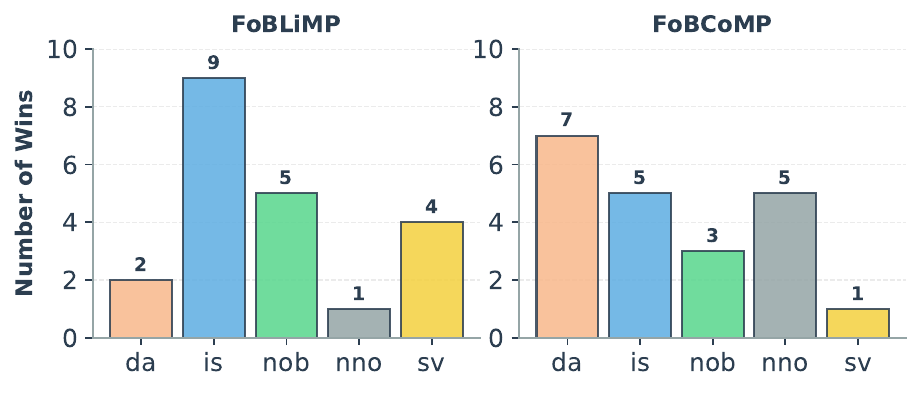}
    \caption{Win rates of transfer languages across models and setups. In case of a tie, we count both.}
    \label{fig:win-la}
\end{figure}

Across most benchmarks, initializing Faroese models with a Scandinavian transfer language improves performance compared to English-only models. The choice of transfer language, however, matters. 

\paragraph{Perplexity} 
Models adapted via a Scandinavian transfer language consistently show lower perplexities than those adapted directly from English. The difference is larger for smaller models and full fine-tuning, which may be a result of larger models and LoRA models reaching lower perplexity than small and fully fine-tuned models. 
Table~\ref{tab:post_adaptation_perplexities_main} shows that after adaptation on Faroese, the best individual transfer language varies between Icelandic and the higher-resource Mainland Scandinavian languages. Norwegian-Nynorsk, the lowest-resource Mainland Scandinavian language, consistently has the highest perplexities. We speculate that the models benefit either from morphological similarity to Icelandic or from the quality and diversity of higher-resource corpora, both of which Nynorsk lacks. 

\paragraph{FoBLiMP}
While Icelandic does not always yield the best language modeling performance (measured in perplexity after Faroese CPT), it achieves the highest MultiBLiMP scores among the individual transfer languages in 3 of 4 setups, as expected since Icelandic has subject-verb agreement, unlike the mainland Scandinavian languages.
Across all linguistic probes, Icelandic is the most frequent winner (Figure~\ref{fig:win-la}; 9 of 20 task-setup pairs, ahead of Bokm\aa l with 5 and Swedish with 4). Mean scores across FoBLiMP tasks (Table~\ref{tab:mean_foblimp}) show Icelandic performs best in 2 of 4 setups (both 360M models), while Swedish is best for both 135M models, making Icelandic the best individual language but not an undisputed leader. 
Low-resource Nynorsk performs worst, with a single win (Figure~\ref{fig:win-la}) and the lowest mean score in all four setups (Table~\ref{tab:mean_foblimp}). 
This suggests that other features can compensate for lower similarity: Swedish matches Icelandic for the smaller models (except in SVA).

\paragraph{FoBCoMP} 
Danish leads with 7 wins. Icelandic remains a strong transfer language, but Nynorsk has the same win count (5), despite poor results in linguistic probes. Four of Nynorsk's wins come from the sentiment tasks with 135M models, while Nynorsk models otherwise perform worse, as reflected in the mean scores over FoBCoMP tasks, where Nynorsk is lowest in 2 of 4 setups (Table~\ref{tab:focomp-mean}). In the mean scores in Table~\ref{tab:focomp-mean}, we also see that Icelandic and Danish both win 2 out of 4 setups. 

\paragraph{Discussion}

Icelandic is the strongest transfer language overall, yielding the lowest perplexity prior to Faroese CPT and clear advantages on FoBLiMP probes, especially subject--verb agreement. 
Compared to English-only initialization, Icelandic improves FoBLiMP in 17/20 cases and FoBCoMP in 11/20 cases. 
On FoBLiMP, all five transfer languages outperform English in 9/20 cases (5/10 for 135M, 4/10 for 360M). For FoBCoMP, transfer provides less consistent gains, with all five transfer languages beating English in only 5/20 cases. 
Across individual transfer languages over both benchmarks (Figure~\ref{fig:win-la}; results of both sides added), Icelandic leads with 14 wins vs.\ 9 for Danish, 8 for Bokm\aa l, 6 for Nynorsk and 5 for Swedish. In direct comparison, Icelandic beats Danish on FoBLiMP (12 vs.\ 6; 2 ties), while Danish performs slightly better on FoBCoMP (11 vs.\ 9).\footnote{In experiments on English$\to$Faroese translation in Appendix~\ref{app:mt}, we find that Swedish is the strongest transfer language; however, there, similarity to English may play a role. }

The overall finding that initializing Faroese models with a Scandinavian transfer language improves performance compared to English-only models confirms prior results for Faroese encoder models \citep{snaebjarnarson-etal-2023-transfer} and extends them to decoder models. \citet{snaebjarnarson-etal-2023-transfer} report that an Icelandic BERT base model often outperforms a Danish base model, consistent with our results for perplexity and linguistic acceptability. Interestingly, they find that Danish BERT performs better on a semantic text similarity benchmark (the only semantic evaluation task in their work), aligning with our observation that Danish performs slightly better than Icelandic on comprehension tasks. More broadly, our findings are consistent with prior work on language transfer, highlighting the benefits of using typologically related languages \citep{chronopoulou-etal-2023-language, faisal-anastasopoulos-2022-phylogeny, chang-etal-2024-multilinguality}.

\subsection{Effects of Merging}
\label{secsec:rq2}

\paragraph{Perplexities} 
We see in Table~\ref{tab:post_adaptation_perplexities_main} that merging leads to the lowest perplexities in full-parameter CPT, suggesting that with sufficient learning capacity, models can benefit from a mixture of related languages. Results for the different merging setups barely differ, showing that results are robust to language weighting and composition. 
In contrast to full fine-tuning, LoRA merges generally underperform individual languages with the only exception of the worst individual language, Nynorsk. For LoRA, the multilingual CPT baseline gets the lowest perplexities, showing that it is not interference of the languages themselves but merging itself that is causing the problem.
Full fine-tuning thus appears more robust to language merging; here, it outperforms multilingual CPT. Merging LoRA-fine-tuned models may generally be more problematic than merging fully fine-tuned models, as other work has also shown, possibly due to a weaker degree of representation alignment \citep{stoica2024modelmergingsvdtie}. 

\paragraph{FoBLiMP}
In Table~\ref{tab:mean_foblimp}, we see that merges are beneficial for the full tuning setup: For the 135M model, the two best models are merges (Merge$^{eq}$ and Merge$^{is+}$), closely followed by Swedish. For the 360M model, Icelandic is best, but followed by all 3 merges. For LoRA however, the situation is different: For the 135M model, merges are within the same range as models with individual languages, while for the 360M model, the scores of merges are lower than for 4 out of 5 languages.

\paragraph{FoBCoMP}
The Merge$^{da+is}$ setup, which combines Icelandic and Danish equally, is the best merge in 3 of 4 setups in Table~\ref{tab:focomp-mean}. Full fine-tuning benefits merges in particular: Merge$^{da+is}$ is the best overall setup for 135M, and Merge$^{is+}$is the best setup for 360M. For LoRA however, single-language setups outperform merges, which mostly do not even surpass English-only. 
Merge$^{da+is}$ is better than Icelandic only for both full fine-tuning models, while Icelandic only is better for both LoRA models (by a small margin for 135M), again demonstrating that merges show promise for full fine-tuning but less so for LoRAs.

\paragraph{Human Ratings} in the sentence continuation task (Table~\ref{tab:all-scores_sentence}) show that the models trained on Icelandic only and the Merge$^{is+}$ model are very close: For LoRA, Merge$^{is+}$ achieves the highest overall score (3.436 versus 3.396), while for full tuning, the Icelandic model achieves higher average scores (3.413 versus 3.361). Icelandic LoRA performs best on the lexical level but is weaker in grammar, semantics and fluency. For semantics, the merged models score slightly higher for LoRA, while both transfer setups perform similarly for full fine-tuning. 
This indicates that while lexical knowledge can be effectively acquired through pre-training on Icelandic, other skills can benefit from the broader exposure provided by the merging approach. 
In the summarization task however (Table~\ref{tab:summ_performance}), scores of the Icelandic model are higher in both cases and across both linguistic quality and task completion, which could indicate the opposite. Fully fine-tuned models here clearly outperform LoRA models on summarization, on both criteria. 

\begin{table*}[t]
    \centering
    \scriptsize
    \begin{minipage}{0.65\textwidth}
        \centering
        \adjustbox{max width=\linewidth}{%
        \begin{tabular}{l ccccc}
        \toprule
         & Overall & Lex. & Gram. & Sem. & Fluency \\\midrule
        LoRA-Merge$^{is+}$   & \bfseries 3.436 & 3.865 & \bfseries 3.645 & \bfseries 3.015 & \bfseries 3.216 \\
        LoRA-Is             & 3.396 & \bfseries 3.983 & 3.587 & 2.891 & 3.126 \\
        Full-Is             & 3.413 & 3.940 & 3.600 & 2.955 & 3.160 \\
        Full-Merge$^{is+}$   & 3.361 & 3.924 & 3.535 & 2.934 & 3.051 \\
        \bottomrule
        \end{tabular}%
        }
        \subcaption{Sentence Continuation. Pearson $r=0.546$.}
        \label{tab:all-scores_sentence}
    \end{minipage}%
    \begin{minipage}{0.31\textwidth}
        \centering
        \adjustbox{max width=\linewidth}{%
        \begin{tabular}{cc}
        \toprule
         Task Compl. & Ling. Quality \\\midrule
        1.03 & 1.78 \\
        1.20 & 2.02 \\
        \bfseries 3.01 & \bfseries 4.11 \\
        2.92 & 3.82 \\
        \bottomrule
        \end{tabular}%
        }
        \subcaption{Summarization. $r = 0.879$.}
        \label{tab:summ_performance}
    \end{minipage}
    \caption{Results for the human evaluation. Scores are averages over annotators. Scale 0--5, higher is better. }
    \label{tab:combined_evaluation}
\end{table*}

\paragraph{Discussion} 

Merging models fine-tuned on related languages can meaningfully benefit CPT in a new target language in the full fine-tuning setup. 
Unlike task merging, where performance on the original tasks is generally preserved or improved \citep{yadav2024mattersmodelmergingscale}, previous attempts at language merging have shown that out-of-the-box performance often deteriorates \citep{glocker2025growmergescalingstrategies}.\footnote{Our out-of-the-box perplexities on Faroese FineWeb confirm this trend, with merged models performing worse than individual source languages; see Table~\ref{tab:base_perplexities} in Appendix~\ref{app:zero-shop-ppl}.} 
In contrast, by continuing pre-training on the target language after merging, we observe gains especially in perplexity, demonstrating that while it may not lead to strong performance out-of-the-box, language merging is a promising \textit{initialization} strategy for further fine-tuning, as the latter apparently helps resolve the parameter interference introduced by merging.

Our results confirm the observation from prior work that LoRAs are more challenging to merge than fully fine-tuned models \citep{stoica2024modelmergingsvdtie, tang2024parameterefficient, zhang-zhou-2025-unraveling}; almost across the board, LoRA merges underperform. 

Finally, although prior work indicates that larger models are easier to merge \citep{dang2024ayaexpansecombiningresearch, yadav2024mattersmodelmergingscale}, we do not observe a strong size effect in our CPT settings. This however does not preclude effects for substantially larger models, where merging may become even more impactful. 

\subsection{Qualitative Observations}

The low semantic scores across models confirm a limitation in small models, especially for low-resource languages like Faroese. While the models acquire competency in surface-level linguistics --- producing valid Faroese vocabulary, maintaining grammatical structures, and achieving natural-sounding fluency --- they struggle significantly with generating meaningful, coherent content. 
This reflects a fundamental challenge in language model training: while syntactic and lexical patterns can be learned from limited data through transfer from related languages, semantic understanding, which requires world knowledge, is a challenge. 

Many summarization outputs exhibit mixes between languages, as in the example in Figure~\ref{fig:example_summary}. This was not the case for sentence continuation outputs, which indicates that the low-resource tuning interferes with the models' linguistic abilities. 

\begin{figure}
    \centering
\begin{tcolorbox}[colback=white,
                  colframe=black,
                  boxrule=0.3pt,
                  arc=1.5mm,
                  left=1mm,
                  right=1mm,
                  top=1mm,
                  bottom=1mm]

\footnotesize
\textbf{Example of summary output:}\\[0.5ex]
Samandráttur er ein \textit{lokaliserendre i} \ul{fjórðhálsparafjöllum fyrir fjölkvangna forfælja í Føroyskaflokkum} til víkjandi barna uppaling er lutfalsliga sterk í Føroyum.
\end{tcolorbox}
\vspace{-10pt}
    \caption{Language mixing in outputs. The model starts in Faroese, then uses language similar to Danish or Norwegian (in italics), then language resembling Icelandic (underlined), then, again, Faroese. }
    \label{fig:example_summary}
\end{figure}

\subsection{Generality of Results}
To assess generalizability, we replicate our approach for Northern Sámi (Uralic) and Upper Sorbian (West Slavic), both even lower-resourced than Faroese and less related to English, the language of the base model. The core findings hold: typologically more related languages (Finnish/Estonian for Northern Sámi; Czech/Polish for Upper Sorbian) consistently outperform majority/contact languages (Swedish/Norwegian; German) on perplexity, confirming our finding from Section~\ref{secsec:rq1} about typological similarity as an important factor for transfer effectiveness. As in Section~\ref{secsec:rq2}, merged models under full fine-tuning again achieve the lowest perplexities, confirming merging as a particularly successful initialization strategy for further continued pre-training on the target language. 
Details are reported in Appendix~\ref{app:hsb_se}. 

\section{Conclusion}

This article examined strategies for adapting small LLMs to Faroese, comparing transfer from related Scandinavian languages and model merging across multiple source languages.
Our results show that transfer from related languages is beneficial, improving perplexity in all settings and linguistic acceptability in most, but no single source language dominates: Icelandic helped most for linguistic accuracy, while Danish was slightly better for comprehension.
This lack of a single best source language suggests drawing on multiple source languages simultaneously, which motivated our experiments on model merging and multilingual continued pre-training.
Under full fine-tuning, merging consistently improved perplexity over the best single source language, outperformed joint multilingual pre-training, and yielded the best comprehension scores for both model sizes, showing that merged models are a successful initialization strategy for continued pre-training on Faroese.
Merged LoRA adapters, however, underperformed individual source languages. In this setting, joint multilingual fine-tuning was the best initialization, so drawing on multiple source languages is still beneficial; merging simply causes stronger interference for LoRA fine-tunes.
Future work may therefore examine whether adaptive or data-driven merging strategies can yield transfer gains for LoRA as well.

\section*{Limitations}

The scarcity of Faroese evaluation data is a fundamental limitation, both for this study and for low-resource NLP more broadly. Even our human evaluation covers relatively few samples, as it is labor-intensive and costly. Summarization results in particular should be interpreted with caution: the small validation set was not a reliable signal for hyperparameter tuning, and in particular the LoRA models, which are more sensitive to hyperparameters, may not have been trained optimally.

We experiment with only two model sizes, one LoRA configuration, and restrict ourselves to English-only base models, chosen as the cleanest experimental baseline given that there were no small ($<$1B) multilingual models with well-documented training data composition available at the time of writing. Extending this work to multilingual base models that include (some of) the target languages would be a valuable next step.

The English-centric tokenizer of our base models is another limitation: it increases inference cost for non-English languages and may also affect model performance. For real-world deployment, transferring to a tokenizer that includes the Northern Germanic languages in training would be better.

\section*{Acknowledgments}
This research was supported by TrustLLM funded by Horizon Europe GA 101135671 and by the Wallenberg AI, Autonomous Systems and Software Program (WASP) funded by the Knut and Alice Wallenberg Foundation. The computations were enabled by the Berzelius resource provided by the Knut and Alice Wallenberg Foundation at the National Supercomputer Centre and by the National Academic Infrastructure for Supercomputing in Sweden (NAISS), partially funded by the Swedish Research Council through grant agreement no. 2022-06725. 

\bibliography{custom}

\appendix

\section{Training Details}
\label{app:a}

This section provides additional details, supplementing Section~\ref{secsec:training} and Section~\ref{subsec:humeval}. Hyperparameters for continued pre-training and for summarization fine-tuning are reported in Table~\ref{tab:hyperparams}.

\begin{table*}[h!]
\centering
\small
\begin{tabular}{@{}p{2.8cm} p{6.2cm} p{6.2cm}@{}}
\toprule
\textbf{Parameter} & \textbf{Continued pre-training} & \textbf{Summarization fine-tuning} \\
\midrule
Optimizer & AdamW \citep{loshchilov2018decoupled} & AdamW \citep{loshchilov2018decoupled} \\
Context window & 2048 tokens & 2048 tokens \\
Scheduler & Cosine with 5\% warmup & Cosine with 0.1 warmup \newline (also tested: no scheduler) \\
Batch size & 256 (effective) & 8 \\
Learning rate & Full fine-tuning: $5\times10^{-4}$ \newline LoRA: $8\times10^{-4}$ \newline (also tested: $5\times10^{-5}$--$1\times10^{-3}$) & $5\times10^{-5}$ \newline (also tested: $5\times10^{-4}$--$1\times10^{-3}$) \\
Dropout & 0 & 0.1 (also tested: 0) \\
LoRA rank & 256 & 16 (also tested: 8) \\
LoRA $\alpha$ & 512 & 32 (also tested: 16) \\
LoRA modules & Feed-forward (following \citet{kunz2025train}) & -- \\
Training epochs & 5 (Faroese corpus), 1 (all other languages) & 50 (full dataset, 151 samples) \\
Tuning split & -- & 135 training / 16 validation samples \\
Prompting & -- & Faroese, minimalistic setup indicating start of text and summary \\
Hardware & 1 node, 4 or 8 A100 40GB GPUs & 1 A100 40GB GPU \\
Total compute & 5,000 A100 (40GB) hours & -- \\
\bottomrule
\end{tabular}
\caption{Training hyperparameters and setup for continued pre-training and for the summarization fine-tuning experiments in the human evaluation.}
\label{tab:hyperparams}
\end{table*}


\section{Dataset Statistics}
\label{app:dataset_statistics}

This section provides overviews of our two benchmark suites introduced in Section~\ref{subsec:autoeval}. FoBLiMP is described in Table~\ref{tab:foblimp_overview}; FoBCoMP in Table~\ref{tab:fobcomp_overview}. Both benchmarks are publicly released on HuggingFace under identifiers \href{https://huggingface.co/datasets/jekunz/FoBLiMP}{jekunz/FoBLiMP} and \href{https://huggingface.co/datasets/jekunz/FoBCoMP}{jekunz/FoBCoMP}, licensed under Creative Commons licenses: FoBLiMP under Creative Commons Attribution ShareAlike 4.0, FoBCoMP under Creative Commons Attribution Non-Commercial ShareAlike 4.0; consistent with the source datasets. 

 
\begin{table*}[h!]{}
    \centering
    \footnotesize
    \resizebox{0.99\textwidth}{!}{%
    \begin{tabular}{l p{2.2cm} p{2.8cm} p{2.0cm} p{1.4cm} p{2.8cm}}
        \toprule
        & Corruption Type & Source & Subset & Pairs & Corruption Method \\
        \midrule
        \multirow{4}{*}{\rotatebox{90}{-- Existing Work -- -- -- }} 
            & Subject-Verb Agreement & MultiBLiMP \citep{jumelet2025multiblimp10massivelymultilingual} & Full & 232 & \textbf{Automatic} using morphological annotation \\
            & Verb Placement & GermDetect \citep{michael-horbach-2025-germdetect} & All corrupted samples & 2{,}026 & \textbf{Automatic} using dependency trees \\
            & Flipped Words & ScaLA \citep{nielsen-2023-scandeval} & \textit{flip\_neighbors} & 552 & \textbf{Automatic} using simple heuristics \\
            & Deleted Words & ScaLA \citep{nielsen-2023-scandeval} & \textit{delete} & 601 & \textbf{Automatic} using simple heuristics \\
        \midrule
        \multirow{1}{*}{\rotatebox{90}{Ours --}} 
            & Translation Pairs & New & Full & 680 (from 100 source sentences) & \textbf{Machine-translated} sentences ranked after \textbf{human ratings} \\
        \bottomrule
    \end{tabular}%
    }
    \caption{Datasets and corruption methods used for the linguistic acceptability probes included in \textbf{FoBLiMP}.}
    \label{tab:foblimp_overview}
\end{table*}
\begin{table*}[h!]{}
    \centering
    \footnotesize
    \renewcommand{\arraystretch}{1.25} 
    \resizebox{0.99\textwidth}{!}{%
    \begin{tabular}{p{1.8cm} p{2.2cm} p{2.6cm} p{2.0cm} p{4cm}}
        \toprule
        Task & Source & Subset & Pairs & MinPair Generation \\
        \midrule
        Sentiment Classification & FoSent \citep{debess-etal-2024-good} & Sentence; both annotators agree & 91 & Add a sentence at the end: \textit{This is good / bad.}  \\
        Sentiment Classification & FoSent \citep{debess-etal-2024-good} & Document; both annotators agree & 84 & Add a sentence at the end: \textit{This is good / bad.} \\
        \midrule
        Topic Classification & FoSent \citep{debess-etal-2024-good} & Deemed relevant by human annotator & 234 & Add a sentence at the end: \textit{The topic of this article is: X}. Exchange related topics (e.g., \textit{Local News} -- \textit{International News}). \\
        \midrule
        Extractive QA & FoQA \citep{simonsen-etal-2025-foqa} & Full & 21{,}867 (2{,}000 unique texts \& correct answers) & Dataset shuffling (negative answer passage is also part of the context, length controlled) \\
        Extractive QA & FoQA \citep{simonsen-etal-2025-foqa} & Full & 2{,}000 & GPT-4-generated: extract a different, wrong answer with the same length. \\
        \bottomrule
    \end{tabular}%
    }
    \renewcommand{\arraystretch}{1.0} 
    \caption{Datasets and corruption methods used for the text comprehension probes included in \textbf{FoBCoMP}.}
    \label{tab:fobcomp_overview}
\end{table*}

\section{Inter-Annotator Agreement}
\label{app:iaa}

We provide the inter-annotator agreement scores for our human evaluations in Table~\ref{tab:iaa}. 

\begin{table}[ht]
\centering
\adjustbox{max width=0.48\textwidth}{
\begin{tabular}{llccc}
\toprule
\textbf{Task} & \textbf{Dimension} & \textbf{$r$} & \textbf{$\kappa_w$} & \textbf{95\% CI} \\
\midrule
\multirow{5}{*}{Sent.\ Cont.}
  & Sem.\ coh. & 0.427 & 0.383 & [0.303, 0.460] \\
  & Lex.\ cor. & 0.409 & 0.286 & [0.212, 0.357] \\
  & Grammar    & 0.375 & 0.367 & [0.277, 0.449] \\
  & Fluency    & 0.298 & 0.295 & [0.205, 0.379] \\
\midrule
\multirow{3}{*}{Summ.}
  & Task compl.  & 0.728 & 0.718 & [0.644, 0.781] \\
  & Ling.\ qual. & 0.829 & 0.819 & [0.760, 0.866] \\
\bottomrule
\end{tabular}
}
\caption{Inter-annotator agreement: Pearson's $r$ and quadratic-weighted Cohen's $\kappa_w$ with bootstrapped 95\% confidence intervals ($10{,}000$ iterations). 
}
\label{tab:iaa}
\end{table}

\section{Machine Translation}
\label{app:mt}

We additionally evaluate English$\to$Faroese machine translation as a generation task. Models are prompted with five fixed examples sampled from the FLORES+ dev split. We evaluate on FLORES+ devtest and the BOUQuET test set and report corpus-level chrF++ (Table~\ref{tab:mt}).

\begin{table*}[ht]
\centering
\begin{subtable}[t]{0.45\textwidth}
\centering
\ptabL{
En & 12.98 & 21.86 & 13.82 & 24.50 \\\midrule
+Da & 15.97 & 27.32 & 16.47 & 28.92 \\
+Is & 14.74 & 28.61 & 14.89 & 26.00 \\
+No$^B$ & 14.21 & 24.62 & 15.41 & 28.51 \\
+No$^N$ & 13.55 & 25.30 & 14.82 & 27.16 \\
+Sv & \textbf{16.23} & 28.48 & \textbf{19.57} & \textbf{30.77} \\\midrule
Merge$^{eq}$ & 15.07 & 29.21 & 12.99 & 24.65 \\
Merge$^{is+}$ & 14.84 & \textbf{29.84} & 15.65 & 27.19 \\
Merge$^{da+is}$ & 15.38 & 28.80 & 15.43 & 26.32 \\\midrule
Multilingual & 13.66 & 24.08 & 19.29 & 30.04 \\
}
\caption{FLORES+ (devtest). \\ Apertus-8B: 40.20. Median SD: 0.19}
\label{tab:mt-flores-chrf}
\end{subtable}\hfill
\begin{subtable}[t]{0.35\textwidth}
\centering
\ptab{
10.49 & 22.07 & 12.47 & 25.16 \\\midrule
14.91 & 27.70 & 16.95 & 29.88 \\
12.24 & 28.16 & 13.13 & 27.21 \\
12.91 & 25.21 & 14.65 & 29.37 \\
11.75 & 25.56 & 13.73 & 26.92 \\
\textbf{15.07} & 28.66 & \textbf{20.08} & \textbf{30.88} \\\midrule
13.50 & 29.56 & 11.75 & 26.49 \\
13.39 & \textbf{30.27} & 15.02 & 28.16 \\
14.82 & 29.09 & 14.81 & 26.69 \\\midrule
11.82 & 24.65 & 18.55 & 30.48 \\
}
\caption{BOUQuET (test). \\  Apertus-8B: 36.44. Median SD: 0.23}
\label{tab:mt-bouquet-chrf}
\end{subtable}\hfill\mbox{}
\caption{Five-shot English$\to$Faroese machine translation: corpus-level chrF++. Mean over 3 seeds. }
\label{tab:mt}
\end{table*}

All individual transfer languages improve over English-only initialization in every setup on both datasets; the only model below English-only is the LoRA Merge$^{eq}$ adapter for 135M. Swedish is the best individual transfer language, in 3 of 4 setups on FLORES+ and in all 4 on BOUQuET, while Icelandic is the strongest individual language for the fully fine-tuned 360M model on FLORES+. For the fully fine-tuned 360M model, all three merges outperform every individual language on both datasets, with Merge$^{is+}$ scoring highest overall, whereas LoRA merges fall behind the best individual languages. Under LoRA, the joint multilingual model is the second-best model, behind Swedish, while under full fine-tuning it barely improves over English-only. 

Notably however, scores remain far below those of Apertus-8B. We hypothesize that this is a result of a lack of few-shot performance and English comprehension abilities of our models. 

\section{Extended Perplexity Results}
\label{app:zero-shop-ppl}

\begin{table*}[ht]
\centering
\begin{subtable}[t]{0.45\textwidth}
\centering
\ptabL{
En & 73.61 & 58.27 & 73.61 & 58.27 \\\midrule
+Da & 50.98 & 40.45 & 63.23 & 44.38 \\
+Is & \textbf{38.77} & \textbf{30.09} & \textbf{40.05} & \textbf{30.54} \\
+No$^B$ & 48.59 & 44.81 & 56.25 & 44.51 \\
+No$^N$ & 61.58 & 50.28 & 68.02 & 56.62 \\
+Sv & 69.96 & 57.02 & 78.22 & 61.43 \\\midrule
Merge$^{eq}$ & 103.45 & 42.40 & 182.21 & 338.30 \\
Merge$^{is+}$ & 94.95 & 40.16 & 68.81 & 62.88 \\
Merge$^{da+is}$ & 65.28 & 40.39 & 78.46 & 146.20 \\
}
\caption{\textit{Before} continuing training on Faroese.}
\label{tab:base_perplexities}
\end{subtable}\hfill
\begin{subtable}[t]{0.35\textwidth}
\centering
\ptab{
5.26 & 4.20 & 4.21 & 3.54 \\\midrule
4.51 & 3.74 & 3.74 & \textbf{3.19} \\
4.62 & 3.69 & 3.93 & 3.21 \\
4.51 & 3.89 & 3.76 & 3.20 \\
4.80 & 3.91 & 4.04 & 3.45 \\
4.54 & 3.74 & \textbf{3.72} & \textbf{3.19} \\\midrule
4.50 & 3.65 & 4.02 & 3.46 \\
\textbf{4.49} & \textbf{3.63} & 3.96 & 3.37 \\
\textbf{4.49} & 3.64 & 3.97 & 3.40 \\
}
\caption{\textit{After} continuing training on Faroese.}
\label{tab:post_adaptation_perplexities}
\end{subtable}\hfill\mbox{}
\caption{Average per-token perplexity on the FineWeb-2 evaluation set, extending results in Table~\ref{tab:post_adaptation_perplexities_main}.}
\label{tab:perplexity}
\end{table*}

Table~\ref{tab:base_perplexities} extends the post-adaptation perplexity results in Table~\ref{tab:post_adaptation_perplexities_main} by reporting zero-shot perplexities; that is, perplexities \textit{before} Faroese adaptation. Among individual languages, Icelandic performs best, followed by Danish and Norwegian Bokmål; Swedish is only marginally better than English alone under full fine-tuning, and the Swedish LoRA adapters are even worse than English.
\textit{After} Faroese adaptation (Table~\ref{tab:post_adaptation_perplexities}), the ranking changes considerably. Icelandic remains the strongest individual transfer language only for the fully fine-tuned 360M model; for LoRA 360M, Danish and Swedish are marginally better. For the smaller 135M models, higher-resource mainland Scandinavian languages outperform Icelandic, with the exception of Norwegian Nynorsk, the lowest-resource language in our set. Swedish, close to English before adaptation, is among the best individual languages after adaptation in both LoRA setups. Similarly, merged models exhibit very high zero-shot perplexity, yet under full fine-tuning they achieve the lowest perplexities after Faroese adaptation. Zero-shot perplexity is thus not a reliable predictor of post-adaptation performance.

\section{Generality of the Results: Experiments on Northern Sámi and Upper Sorbian}
\label{app:hsb_se}

To assess which of our Faroese findings generalize to other languages, we replicate the approach for two additional low-resource languages: Northern Sámi and Upper Sorbian. Both have fewer speakers and smaller corpora than Faroese and are classified by UNESCO as endangered (Northern Sámi) and vulnerable (Upper Sorbian) \citep{moseley2010atlas}. Neither functions as the majority language in its region, and public life is dominated by majority languages \citep{dagsvold2015can, mcmonagle2022aspects, Mitschke20257795}. 
Typologically, Upper Sorbian is Indo-European (West Slavic), while Northern Sámi belongs to the Uralic family. This allows us to test whether the Faroese findings generalize both within Indo-European and across language families.\footnote{Unlike the Faroese–Icelandic case, the closest transfer languages for Northern Sámi (Finnish) and Upper Sorbian (Czech and Polish) are not themselves low-resource, so we cannot test whether a more distant but higher-resource language improves semantic performance.}

\subsection{Experimental Setup}

We focus on the 360M parameter models and compare full fine-tuning and LoRA using the same hyperparameters as in the Faroese experiments. For each target language and adaptation method, we create two merges: one combining two languages (a typologically close language and the regional majority language), and one including one or two additional related languages.
For Northern Sámi, we use Finnish and Estonian as the related languages \citep{kahn2017north}, and Swedish and Norwegian (Bokmål) as majority, high-contact languages. Most Northern Sámi speakers are bilingual in one of these languages \citep{kahn2017north}, which influences local varieties \citep{jokinen2016variation} and results in lexical borrowing \citep{pietikainen2008sami, pietikainen2010sami}. 
For Upper Sorbian, we select Czech and Polish as West Slavic neighbors and German as the regional majority language; all Sorbs speak German, often as their dominant language \citep{Mitschke20257795}.

\textbf{Evaluation} relies on massively multilingual benchmarks due to the lack of suitable curated resources. We report perplexity on the respective FineWeb-2 test sets and evaluate linguistic acceptability with MultiBLiMP \citep{jumelet2025multiblimp10massivelymultilingual} (2536 pairs for Northern Sámi, 186 for Upper Sorbian) and reading comprehension with MultiWikiQA \citep{smart2025multiwikiqareadingcomprehensionbenchmark} (4146 examples for Northern Sámi, 4906 for Upper Sorbian). For MultiWikiQA, we follow the non-LLM negative sampling procedure used in the Faroese experiments.

\subsection{Results and Conclusions}

The results are reported in Table~\ref{tab:crosslingual_full_results}. 

\begin{table*}[ht]
\centering
\adjustbox{max width=\textwidth}{
\begin{tabular}{llcccccc}\toprule
& & \multicolumn{2}{c}{Perplexity $\downarrow$} & \multicolumn{2}{c}{MultiBLiMP $\uparrow$} & \multicolumn{2}{c}{MultiWikiQA $\uparrow$} \\
\cmidrule(lr){3-4}\cmidrule(lr){5-6}\cmidrule(lr){7-8}
& & Full & LoRA & Full & LoRA & Full & LoRA \\\midrule
\multirow{7}{*}{Se}
& En & 6.52 & 6.80 & 80.52 & 81.11 & \textbf{77.86} & 77.42 \\\cmidrule(lr){2-8}
& +Fi & 6.09 & \textbf{5.15} & 82.22 & \textbf{87.22} & 77.35 & 77.26 \\
& +Ee & 6.25 & 5.42 & 82.61 & 86.00 & 76.46 & \textbf{78.20} \\
& +No$^B$ & 6.27 & 5.22 & 79.73 & 83.24 & 77.42 & 77.35 \\
& +Sv & 6.39 & 5.45 & 79.93 & 83.60 & 76.70 & 78.00 \\\cmidrule(lr){2-8}
& Merge$^{fi+sv}$ & 6.06 & 5.57 & \textbf{82.85} & 83.91 & 77.52 & 76.56 \\
& Merge$^{all}$ & \textbf{5.90} & 5.86 & 82.06 & 83.24 & 77.42 & 76.77 \\\midrule
\multirow{6}{*}{Hsb}
& En & 7.40 & 8.53 & 73.66 & 75.27 & 77.90 & 77.29 \\\cmidrule(lr){2-8}
& +Cs & 6.44 & \textbf{5.87} & \textbf{80.11} & 82.80 & \textbf{78.31} & \textbf{79.27} \\
& +De & 6.96 & 6.55 & 72.58 & 79.03 & 76.50 & 77.25 \\
& +Pl & 6.55 & 5.89 & 75.27 & \textbf{84.41} & 78.11 & 79.15 \\\cmidrule(lr){2-8}
& Merge$^{de+pl}$ & 6.49 & 6.56 & 75.27 & 79.57 & 76.44 & 77.07 \\
& Merge$^{all}$ & \textbf{6.33} & 6.33 & 75.81 & 82.80 & 77.48 & 77.29 \\\bottomrule
\end{tabular}
}
\caption{Results for Northern Sámi (Se) and Upper Sorbian (Hsb) with 360M models: per-token perplexity on FineWeb-2, and percentage of correctly ranked pairs on MultiBLiMP and MultiWikiQA.}
\label{tab:crosslingual_full_results}
\end{table*}

\paragraph{Effect of transfer languages.}
A first continued pre-training step on a related language consistently improves perplexity compared to models without a transfer language. In both languages, typologically related transfer languages yield the largest gains: Finnish and Estonian outperform Norwegian Bokmål and Swedish for Northern Sámi, and Czech and Polish outperform German for Upper Sorbian. Majority and contact languages also help, but generally less so. For Faroese, the typologically closest language (Icelandic) yielded the lowest perplexity only in one of four setups, with higher-resource Mainland Scandinavian languages being equally or more effective; since the closest transfer languages for Northern Sámi and Upper Sorbian are not themselves low-resource, typological similarity and data size do not compete here.
For linguistic acceptability probes (MultiBLiMP), typologically closer languages consistently improve over the English-only baseline. For more distant contact languages, LoRA models show consistent improvements over English while fully fine-tuned models do not, suggesting that stronger regularization helps when transfer languages are less related.
Results on comprehension probes are less clear. For Northern Sámi, MultiWikiQA differences are marginal across setups. Upper Sorbian shows somewhat clearer improvements from related-language transfer, though gains remain small overall.
Overall, these results confirm our Faroese findings from Section~\ref{secsec:rq1}: transfer languages help, gains on comprehension probes are less consistent than on linguistic acceptability, and typological, in particular morphological, similarity is important for linguistic acceptability.

\paragraph{Effect of merging.}
Under full fine-tuning, merged models again achieve the lowest perplexities, confirming our finding from Section~\ref{secsec:rq2} that combining multiple transfer sources improves language modeling. Merged LoRA adapters, however, underperform the best single-language variants, as for Faroese.
For linguistic acceptability, results are mixed. For Northern Sámi MultiBLiMP under full fine-tuning, the Finnish--Swedish merge scores highest, while for Upper Sorbian a single language (Czech) wins. Under LoRA, a single language always scores highest (Finnish and Polish, respectively). For MultiWikiQA, merges and individual languages perform similarly for Northern Sámi under full fine-tuning, while merged LoRA adapters score lowest and Czech is the consistent winner for Upper Sorbian. We conclude that merging reliably helps language modeling and can help linguistic acceptability, but does not help comprehension in these languages. This differs from Faroese, where fully fine-tuned merges achieved the best comprehension scores (Section~\ref{secsec:rq2}), suggesting that this benefit does not generalize across languages.

\paragraph{LoRA versus full fine-tuning.}
As in the Faroese experiments, LoRA often achieves lower perplexity for Northern Sámi and Upper Sorbian. The same holds for linguistic acceptability, where LoRA consistently outperforms full fine-tuning, especially for Upper Sorbian, mirroring the higher FoBLiMP scores of LoRA models for Faroese. This may reflect stronger regularization when adapting to small corpora. QA results remain inconclusive, particularly for Northern Sámi where differences across setups are minimal.

\paragraph{Limitations} For these very low-resource languages, data quality for both training and evaluation data may be poor. For linguistic acceptability and comprehension, we have only one dataset each, which is synthetically generated (although, hopefully, from real texts in the target language\footnote{\say{Hopefully} because, as \citet{tatariya-etal-2026-good} show, Wikipedia can consist of primarily extremely low-quality data for smaller languages. However, Northern Sámi and Upper Sorbian fall under their second best out of four tiers, indicating that these Wikipedias are at least not \textit{extremely} bad.}), making the reliability of the results uncertain.

\end{document}